%% file: main.tex
\documentclass[letterpaper]{article} 
\usepackage{aaai23}  

\usepackage{times}  
\usepackage{helvet}  
\usepackage{courier}  
\usepackage[hyphens]{url}  
\usepackage{graphicx} 
\usepackage{natbib}  
\usepackage{caption} 
\usepackage{algorithm}
\usepackage{algorithmic}
\usepackage{todonotes}
\usepackage{booktabs}
\usepackage{multirow}

\definecolor{beamer}{RGB}{41, 52, 55}

\usepackage{newfloat}
\usepackage{listings}
\DeclareCaptionStyle{ruled}{labelfont=normalfont,labelsep=colon,strut=off} 
\floatstyle{ruled}
\newfloat{listing}{tb}{lst}{}
\floatname{listing}{Listing}
\input{notation.tex}

\usepackage[acronym, nohypertypes={acronym}]{glossaries}

\newacronym{model}{SGP}{\emph{Scalable Graph Predictor}}

\title{Scalable Spatiotemporal Graph Neural Networks}
\author {
    Andrea Cini \textsuperscript{\rm \equalcontrib 1},
    Ivan Marisca \textsuperscript{\rm \equalcontrib 1},
    Filippo Maria Bianchi \textsuperscript{\rm 23},
    Cesare Alippi \textsuperscript{\rm 14}
}
\affiliations {
    \textsuperscript{\rm 1} IDSIA, Universit\`a della Svizzera italiana,
    \textsuperscript{\rm 2} UiT the Arctic University of Norway,\\
    \textsuperscript{\rm 3} NORCE Norwegian Research Centre,
    \textsuperscript{\rm 4} Politecnico di Milano\\[.3em]
    \texttt{andrea.cini@usi.ch, ivan.marisca@usi.ch,\\filippo.m.bianchi@uit.no, cesare.alippi@usi.ch}
}

\begin{document}

\maketitle

\begin{abstract}
Neural forecasting of spatiotemporal time series drives both research and industrial innovation in several relevant application domains. Graph neural networks~(GNNs) are often the core component of the forecasting architecture. However, in most spatiotemporal GNNs, the computational complexity scales up to a quadratic factor with the length of the sequence times the number of links in the graph, hence hindering the application of these models to large graphs and long temporal sequences. While methods to improve scalability have been proposed in the context of static graphs, few research efforts have been devoted to the spatiotemporal case. To fill this gap, we propose a scalable architecture that exploits an efficient encoding of both temporal and spatial dynamics. In particular, we use a randomized recurrent neural network to embed the history of the input time series into high-dimensional state representations encompassing multi-scale temporal dynamics. Such representations are then propagated along the spatial dimension using different powers of the graph adjacency matrix to generate node embeddings characterized by a rich pool of spatiotemporal features. The resulting node embeddings can be efficiently pre-computed in an unsupervised manner, before being fed to a feed-forward decoder that learns to map the multi-scale spatiotemporal representations to predictions. The training procedure can then be parallelized node-wise by sampling the node embeddings without breaking any dependency, thus enabling scalability to large networks.
Empirical results on relevant datasets show that our approach achieves results competitive with the state of the art, while dramatically reducing the computational burden.
\end{abstract}

\section{Introduction}

As graph neural networks~(GNNs;~\citealt{scarselli2008graph, bacciu2020gentle}) are gaining more traction in many application fields, the need for architectures scalable to large graphs -- such as those associated with large sensor networks -- is becoming a pressing issue. While research to improve the scalability of models for static graph signals has been very prolific~\cite{hamilton2017inductive, chiang2019cluster, zeng2019graphsaint, frasca2020sign}, little attention has been paid to the additional challenges encountered when dealing with discrete-time dynamical graphs, i.e., spatiotemporal time series. Several of the existing scalable training techniques rely on subsampling graphs to reduce the computational requirements of the training procedure, e.g.,~\cite{hamilton2017inductive, zeng2019graphsaint}. However, sampling the node-level observations as if they were i.i.d.\ can break relational~(spatial) dependencies in static graphs and it is even more problematic in the dynamic case, as dependencies occur also across the temporal dimension. Indeed, complex temporal and spatial dynamics that emerge from the interactions across the whole graph over a long time horizon, can be easily disrupted by perturbing such spatiotemporal structure with subsampling. As an alternative, precomputing aggregated features over the graph allows for factoring out spatial propagation from the training phase in certain architetures~\cite{frasca2020sign}. However, similarly to the subsampling approach, extending this method to the spatiotemporal case is not trivial as the preprocessing step must account also for the temporal dependencies besides the graph topology.

\begin{figure}[t]
\centering
\includegraphics[width=\columnwidth]{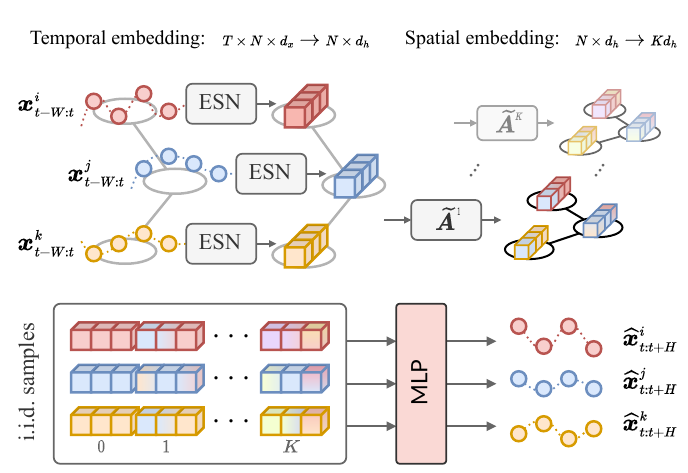}
\caption{Overview of the forecasting framework. Light-grey boxes denote training-free components. At first, an Echo-State Network (ESN) -- with shared parameters across nodes -- encodes multi-scale temporal dynamics. Then, $K$ graph shift operators are used to propagate spatial information. The resulting $K+1$ representations are concatenated and fed to an MLP to predict the next $H$ node observations.}
\label{fig:summary}
\end{figure}

In this paper, we propose a novel scalable encoder-decoder architecture for processing spatiotemporal data; Fig.~\ref{fig:summary} shows high-level schematics of the proposed approach. The spatiotemporal \textit{encoding} scheme is training-free: first, it exploits a deep randomized recurrent neural network~\cite{jaeger2001echo, gallicchio2017deep} to encode the history of each sequence in a high-dimensional vector embedding; then, it uses powers of the graph adjacency matrix to build informative node representations of the spatiotemporal dynamics at different scales. According to the downstream task at hand, the \textit{decoder} maps the node representations into the desired output, e.g., the future values of the time series associated with each node. To improve efficiency, we exploit the structure of the extracted embedding to design the decoder to act as a collection of filters localized at different spatiotemporal scales. 

Since the spatiotemporal encoder requires neither training nor supervision, the representation of each node and time step can be computed in a preprocessing stage, without the constraints that come from online training on GPUs with limited memory. The decoder is the only component of the architecture with trainable parameters.
However, since spatiotemporal relationships are already embedded in the representations, the embeddings can be processed independently from their spatiotemporal context with two consequent advantages. First, training can be done node-wise, allowing for sampling node representations in mini-batches of a size proportional to the hardware capacity.
Second, the decoder can be implemented similarly to a standard multilayer perceptron~(MLP) readout, which is fast and easy to train. 

Let $T$ and $E$ be the number of steps and the number of edges in the input graph, respectively. The cost of training a standard spatiotemporal GNN on a mini-batch of data has a computational and memory cost that scales as $\bigO(TE)$, or $\bigO(T^2E)$ in attention-based architectures~\cite{wu2021traversenet}. Conversely, in our approach mini-batches can be sampled disregarding the length of the sequence and size of the graph, thus making scalability in training constant, i.e., $\bigO(1)$, w.r.t.\ the spatiotemporal dimension of the problem.

Our contributions can be summarized as follows.
\begin{itemize}
    \item We propose a general scalable deep learning framework for spatiotemporal time series, which exploits a novel encoding method based on randomized recurrent components and scalable GNNs architectures.
    \item We apply the proposed model to forecast multivariate time series, whose channels are subject to spatial relationships described by a graph.
    \item We carry out a rigorous and extensive empirical evaluation of the proposed architecture and variations thereof. Notably, we introduce \emph{two} benchmarks for scalable spatiotemporal forecasting architectures.
\end{itemize}
Empirical results show that our approach performs on par with the state of the art while being easy to implement,  computationally efficient, and extremely scalable. Given these considerations, we refer to our architecture as~\textbf{\gls{model}}.

\section{Preliminaries and Problem Definition}~\label{sec:problem}
We consider discrete-time spatiotemporal graphs. In particular, given $N$ interlinked sensors, we indicate with ${\vx_t^i \in \sR^{d_x}}$ the $d_x$-dimensional multivariate observation associated with the $i$-th sensor at time-step $t$, with ${\mX_t \in \sR^{N \times d_x}}$ the node attribute matrix encompassing measurements graph-wise, and with $\mX_{t:t+T}$ the sequence of $T$ measurements collected in the time interval $[t, t+T)$ at each sensor. Similarly, we indicate with ${\mU_t \in \sR^{N\times d_u}}$ the matrix containing exogenous variables~(e.g., weather information related to a monitored area) associated with each sensor at the $t$-th time-step. Then, we indicate additional, optional, static node attributes as ${\mV \in \sR^{N\times d_v}}$. The relational information is encoded in a, potentially dynamic, weighted adjacency matrix ${\mA_t\in\sR^{N\times N}}$. We indicate with the tuple ${\gG_t=\langle\mX_t, \mU_t, \mV, \mA_t \rangle}$ the graph signal at the $t$-th time-step. Note that the number of sensors in a network is here considered fixed only to ease the presentation; we only request nodes to be distinguishable across time steps. The objective of spatiotemporal forecasting is to predict the next $H$ observations given a window of $W$ past measurements. In particular, we consider the family of forecasting models ${F_\theta({}\cdot)}$ s.t.
\begin{equation}
    \widehat \mX_{t:t+H} = F_\theta\left(\gG_{t-W:t}\right),
\end{equation}
where $\theta$ indicates the learnable parameters of the model and $\widehat \mX_{t:t+H}$ the $H$-step ahead point forecast. 

\paragraph{Echo-State Networks} Echo state networks~\cite{jaeger2001echo, lukovsevivcius2009reservoir} are a class of randomized architectures that consist of recurrent neural networks with random connections that encode the history of input signals into a high-dimensional state representation to be used as input to a (trainable) readout layer. The main idea is to feed an input signal into a high-dimensional, randomized, and non-linear reservoir, whose internal state can be used as an embedding of the input dynamics. 
An echo state network is governed by the following state update equation:
\begin{equation}\label{e:esn}
    \vh_t = \sigma\left(\mW_{x} \vx_t + \mW_{h} \vh_{t-1} + \vb\right),
\end{equation}
where $\vx_t$ indicates  a generic input to the system, $\mW_x\in \sR^{d_h\times d_x}$ and $\mW_h\in \sR^{d_h\times d_h}$ are the random matrices defining the connectivity pattern in the reservoir, $\vb\in\sR^{d_h}$ is a randomly initialized bias, $\vh_t$ indicates the reservoir state, and $\sigma$ is a nonlinear activation function~(usually~$\text{tanh}$). If the random matrices are defined properly, 
the reservoir will extract a rich pool of dynamics characterizing the system underlying the input time series $\vx_t$ and, thus, the reservoir states become informative embeddings of $\vx_{t-T: t}$~\cite{lukovsevivcius2009reservoir}. Thanks to the non-linearity of the reservoir, the embeddings are commonly processed with a linear readout that is optimized with a least squares procedure to perform classification, clustering, or time series forecasting~\cite{bianchi2020reservoir}.

\begin{figure*}[t]
\centering
\includegraphics[width=0.98\textwidth]{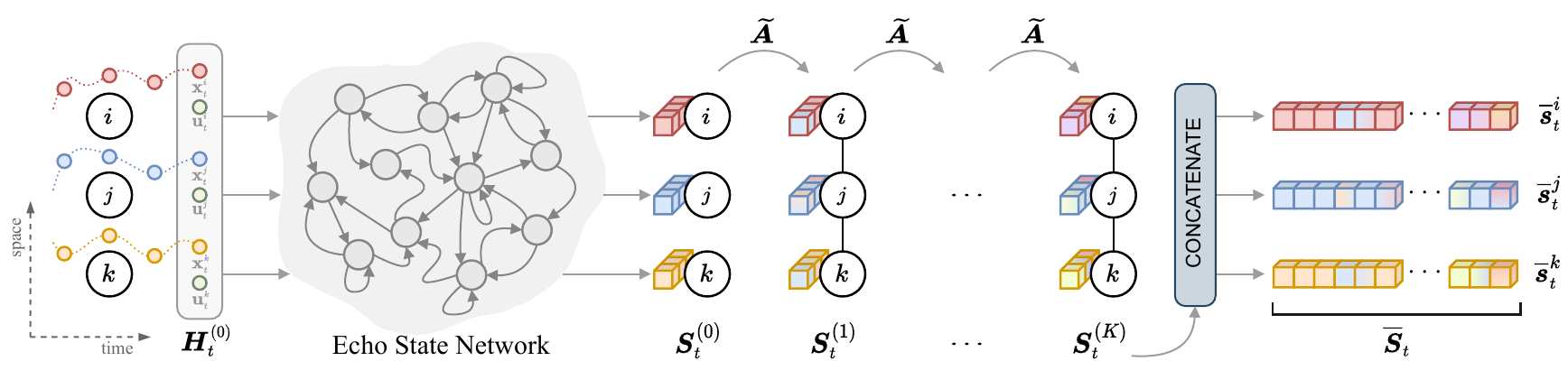}
\caption{Overview of the \gls{model} encoder. Input time series are fed into a randomized network with recurrent connections and embedded into a hierarchical vector representation. A graph shift operator is used to propagate and aggregate spatial information of different order which is then concatenated to obtain a final embedding.}
\label{fig:architecture}
\end{figure*}

\section{Scalable Spatiotemporal GNNs}\label{sec:method}

This section presents our approach to building scalable GNN architectures for time series forecasting. Our method is based on a hybrid encoder-decoder architecture. The encoder first constructs representations of the time series observed at each node by using a reservoir that accounts for dynamics at different time scales. Representations are further processed to account for spatial dynamics described by the graph structure. In particular, as shown on the right-hand side of Fig.~\ref{fig:architecture}, we use incremental powers of the graph adjacency matrix to propagate and aggregate information along the spatial dimension. Each power of the propagation matrix accounts for different scales of spatial dynamics. The final embedding is then built by concatenating representations obtained w.r.t.\ each propagation step, thus resulting in a rich encoding of both spatial and temporal features.

The encoder does not need any training and, once computed, the embeddings can be uniformly sampled over time and space when training a nonlinear readout to perform $H$-step-ahead predictions. The straightforward choice for the decoder~(i.e., readout) is to map the encodings to the outputs~(i.e., predictions) by using a linear transformation or a standard MLP. However, to further enhance scalability, our decoder exploits the structure of the embedding to reduce the number of parameters and learn filters that are localized in time and space. As we will discuss in Sec.~\ref{s:decoder}, this is done by learning separate weight matrices for each spatiotemporal scale.

The following subsections describe in detail each component of the architecture.

\subsection{Spatiotemporal Encoder}\label{s:encoder} 

We consider as temporal encoders deep echo state networks~(DeepESN;~\citealt{gallicchio2017deep}) with leaky integrator neurons~\cite{jaeger2007optimization}. In particular, we consider networks where the signal associated with each node is encoded by a stack of $L$ randomized recurrent layers s.t.\

\begin{align}\label{e:deepesn}
\begin{split}
    \vh^{i,(0)}_t &=\left[\vx_t^i \| \vu_t^i\right],\\
    \hat\vh^{i,(l)}_t &= \text{tanh}\left(\mW^{(l)}_{u} \vh_t^{i,(l-1)} + \mW^{(l)}_{h} \vh_{t-1}^{^{i,(l)}} +\vb^{(l)}\right),\\
    \vh^{i,(l)}_t &= (1 - \gamma_l) \vh_{t-1}^{i,(l)} +\gamma_l \hat\vh^{i,(l)}_t, \qquad l=1,\dots,L
    \end{split}
\end{align}

where $\gamma_l \in \left(0, 1\right]$ is a discount factor associated with $l$-th layer, $\mW^{(l)}_u\in \sR^{d_{h^{l}}\times d_{h^{l-1}}}$, $\mW_h\in \sR^{d_{h^{l}}\times d_{h^{l}}}$, $\vb\in\sR^{d_{h^{l}}}$ are random weight matrices, $\vh^{i,(l)}_t$ indicates the hidden state of the system w.r.t.\ the $i$-th node at the $l$-th layer, and ${}\|{}$ indicates node-wise concatenation. As Eq.~\ref{e:deepesn} shows, DeepESNs are a hierarchical stack of reservoir layers that, e.g., by changing the discount factor at each layer, extract a rich pool of multi-scale temporal dynamics~\cite{gallicchio2017deep}\footnote{We refer to~\cite{gallicchio2018design} for more details on the properties and stability of DeepESNs.}. Given a DeepESN encoder, the input is represented by the concatenation of the states from each layer, i.e., we obtain node-level temporal encodings $\overline\vh_t^i$ for each node $i$ and time-step $t$ as
\begin{equation}\label{e:hbar}
    \overline\vh_t^i = \left(\vh^{i, (0)}_t \| \vh^{i, (1)}_t\| \dots \| \vh_t^{i, (L)}\right).
\end{equation}
We indicate as $\overline\mH_t$ the encoding for the whole graph at time $t$. The extraction of the node-level temporal embeddings is depicted on the left-end side of Fig.~\ref{fig:architecture}, where, to simplify the drawing, we depict an ESN with a single layer.

The next step is to propagate information along the spatial dimension. As discussed at the beginning of the section, we use powers of a graph shift operator $\widetilde\mA$ to propagate and aggregate node representations at different scales. By using a notation similar to Eq.~\ref{e:hbar}, we obtain spatiotemporal encodings as
\begin{align}
    \mS^{(0)}_t &= \overline\mH_t = \left(\mH^{(0)}_t \|\mH^{(1)}_t\| \dots \| \mH^{(L)}_t\right),\notag\\
    \mS^{(k)}_t &= \widetilde\mA\mS^{(k-1)}_t = \left(\widetilde\mA^k\mH^{(0)}_t \|\widetilde\mA^k\mH^{(1)}_t\| \dots \| \widetilde\mA^k\mH^{(L)}_t\right),\notag\\
    \overline{\mS}_t  &= \left(\mS^{(0)}_t\|\mS^{(1)}_t\|\dots\|\mS^{(K)}_t\right),\label{e:esign}
\end{align}
where $\widetilde\mA$ indicates a generic graph shift operator matching the sparsity pattern of the graph adjacency matrix. In practice, by indicating with $\mD$ the graph degree matrix, we use $\widetilde \mA = \mD^{-1}\mA$ in the case of a directed graph and the symmetrically normalized adjacency ${\widetilde \mA = \mD^{-1/2}\mA\mD^{-1/2}}$ in the undirected case. Furthermore, for directed graphs we optionally increase the number of representations to $2K + 1$ to account for bidirectional dynamics, i.e., we repeat the encoding process w.r.t.\ the transpose adjacency matrix similarly to~\cite{li2018diffusion}. Intuitively, each propagation step $\widetilde\mA\mS^{(k-1)}_t$ propagates and aggregates~--~properly weighted~--~features between nodes connected by paths of length $k$ in the graph. As shown in Eq.~\ref{e:esign}, features corresponding to each order $k$ can be computed recursively with $K$ sparse matrix-matrix multiplications~(Fig.~\ref{fig:architecture}).
Alternatively, each matrix $\widetilde\mA^k$ can be precomputed and the computation of the different blocks of matrix $\overline\mS_t$ can be distributed in a parallel fashion as suggested in Fig.~\ref{fig:summary}. Even in the case of extremely large graphs, features $\overline{\mS}_t$ can be computed offline by exploiting distributed computing as they do not need to be loaded on the GPU memory. 

\subsection{Multi-Scale Decoder}\label{s:decoder}

The role of the decoder is that of selecting and weighing from the pool of the (possibly redundant) features extracted by the spatiotemporal encoder and mapping them to the desired output. Representations $\overline\mS_t$ can be fed into an MLP that performs node-wise predictions. Since the representations are large vectors, a na\"ive implementation of the MLP results in many parameters that hinder scalability. Therefore, we replace the first MLP layer with a more efficient implementation that exploits the structure of the embeddings.

As we described in Sec.~\ref{s:encoder}, $\overline\mS_t$ is the concatenation of the representations corresponding to different spatial propagation steps which, in turn, are obtained from the concatenation of multi-scale temporal features. To exploit this structure, we design the first layer of the decoder with a sparse connectivity pattern to learn representations ${\overline{\mZ}}_t$ s.t.\ 
\begin{align}
    \mZ^{(k)}_t 
    &=\sigma\left(\widetilde\mA^k\mH^{(0)}_t \mTheta_k^{(0)} \| \dots \| \widetilde\mA^k\mH^{(L)}_t \mTheta_k^{(L)}\right)\label{e:gcns}\\
    &=\sigma\left(\mS^{(k)}_t\left[\begin{smallmatrix}
  \mTheta_k^{(0)} & & \mathbf{0}\\
  & \ddots &\\ 
  \mathbf{0} & & \mTheta_k^{(L)}
\end{smallmatrix}\right]\right),\label{e:sparse}\\
\overline{\mZ}_t  &= \left(\mZ^{(0)}_t\|\mZ^{(1)}_t\|\dots\|\mZ^{(K)}_t\right),\label{e:zbar}
\end{align}
where $\mTheta_k^{(l)}\in \sR^{d_{h^l}\times d_z}$ are the learnable parameters and $\sigma$ is an activation function.
In practice, representations  $\overline{\mZ}_t$ can be efficiently computed by exploiting grouped $1$-d convolutions~(e.g., see \citealt{krizhevsky2012imagenet}) to parallelize computation on GPUs. In particular, if we indicate the $1$-d grouped convolution operator with $g$ groups and kernel size $r$ as ${}\star_{r,g}{}$, and the collection of the decoder parameters $\mTheta_k^{(l)}$ as $\mTheta$ we can compute $\overline{\mZ}_t$ as
\begin{equation}
    \overline{\mZ}_t = \sigma\left( \bm{\mTheta} \star_{1,g} \overline{
\mS}_t\right),
\end{equation}
with $g=L(K+1)$ in the case of undirected graphs and $g=L(2K+1)$ for the directed case. Besides reducing the number of parameters by a factor of $L(K+1)$, this architecture localizes filters $\mTheta_k^{(L)}$ w.r.t.\ the dynamics of spatial order $k$ and temporal scale $l$. In fact, as highlighted in Eq.~\ref{e:gcns}--\ref{e:zbar}, representation $\overline{\mZ}_t$ can be seen as a concatenation of the results of $L(K+1)$ graph convolutions of different order. 
Finally, the obtained representations are fed into an MLP that predicts the $H$-step-ahead observations as
\begin{equation}
    \widehat\vx^i_{t:t+H} = \texttt{MLP}\left(\overline{\vz}^{i}_t, \vv^i\right),
\end{equation}
where the static node-level attributes $\vv^i$ can also be augmented by concatenating a set of learnable parameters~(i.e., a learnable positional encoding).

\paragraph{Training and sampling} The main improvement introduced by the proposed approach in terms of scalability concerns the training procedure.
Representations $\overline{\mS}_t$ embed both the temporal and spatial relationships among observations over the sensor network.
Consequently, each sample $\overline{\vs}^i_t$ can be processed independently since no further spatiotemporal information needs to be collected.
This allows for training the decoder with SGD by uniformly and independently sampling mini-batches of data points $\overline{\vs}^i_t$.
This is the key property that makes the training procedure extremely scalable and drastically reduces the lower bound on the computational complexity required for the training w.r.t.\ standard spatiotemporal GNN architectures. 

\section{Related works}\label{sec:related}

\input{tab_traffic_std}

Spatiotemporal GNNs are essentially based on the idea of integrating message-passing modules in architectures to process sequential data. Notably, \citet{seo2018structured} and \citet{li2018diffusion} use message-passing to implement gates of recurrent neural networks. \citet{yu2017spatio} and \citet{wu2019graph, wu2020connecting} proposed architectures alternating temporal and spatial convolutions. \citet{wu2021traversenet} and \citet{marisca2022learning}, instead, exploit the attention mechanism to propagate information along both time and space. Modern architectures often combine some type of relational inductive bias, with full Transformer-like attention~\cite{vaswani2017attention} along the spatial dimension~\cite{zheng2020gman, oreshkin2021fc, satorras2022multivariate}, which, however, makes the computation scale quadratically with the number of nodes. \gls{model} falls within the category of \emph{time-then-graph models}, i.e., models where the temporal information is encoded before being propagated along the spatial dimension. \citet{gao2022equivalence} showed that such models can be more expressive than architectures that alternate temporal and spatial processing steps.

Research on scalable models for discrete-time dynamic graphs has been relatively limited. Practitioners have mostly relied on methods developed in the context of static graphs which include node-centric, GraphSAGE-like, approaches~\cite{hamilton2017inductive} or subgraph sampling methods, such as ClusterGCN~\cite{chiang2019cluster} or GraphSAINT~\cite{zeng2019graphsaint}. \citet{wu2020connecting, gandhi2021spatio, wu2021inductive} are examples of such approaches. Among scalable GNNs for static graphs, SIGN~\cite{frasca2020sign} is the approach most related to our method. Like in our approach, SIGN performs spatial propagation as a preprocessing step by using different shift operators to aggregate across different graph neighborhoods, which are then fed to an MLP. However, SIGN is limited to static graphs and propagates raw node-level attributes. Finally, similar to our work, DynGESN~\cite{micheli2022discrete} processes dynamical graphs with a recurrent randomized architecture. However, the architecture in DynGESN is completely randomized, while ours is hybrid as it combines randomized components in the encoder with trainable parameters in the decoder.

\section{Empirical evaluation}\label{sec:exp}

\input{tab_ls}

We empirically evaluate our approach in $2$ different scenarios. In the first one, we compare the performance of our forecasting architecture against state-of-the-art methods on popular, medium-scale, traffic forecasting benchmarks. In the second one, we evaluate the scalability of the proposed method on large-scale spatiotemporal time series datasets by considering two novel benchmarks for load forecasting and PV production prediction. 
We provide an efficient open-source implementation of \gls{model} together with the code to reproduce all the experiments\footnote{\url{https://github.com/Graph-Machine-Learning-Group/sgp}}.

\input{tab_data.tex}

\paragraph{Datasets} In the first experiment we consider the \textbf{METR-LA} and \textbf{PEMS-BAY} datasets~\cite{li2018diffusion}, which are popular medium-sized  benchmarks used in the spatiotemporal forecasting literature. In particular, METR-LA  consists of traffic speed measurements taken every $5$ minutes by $207$ detectors in the Los Angeles County Highway, while PEMS-BAY includes analogous observations recorded by $325$ sensors in the San Francisco Bay Area. We use the same preprocessing steps of previous works to extract a graph and obtain train, validation and test data splits~\cite{wu2019graph}. For the second experiment, we introduce two larger-scale datasets derived from energy analytics data. The first dataset contains data coming from the Irish Commission for Energy Regulation Smart Metering Project~(\textbf{CER-E};~\citealt{cer2016cer}), which has been previously used for benchmarking spatiotemporal imputation methods~\cite{cini2022filling}; however, differently from previous works, we consider the full sensor network consisting of $6435$ smart meters measuring energy consumption every $30$ minutes at both residential and commercial/industrial premises. The second large-scale dataset is obtained from the synthetic \textbf{PV-US}\footnote{\url{https://www.nrel.gov/grid/solar-power-data.html}} dataset~\cite{hummon2012sub}, consisting of simulated energy production by $\text{5016}$ PV farms scattered over the United States given historic weather data for the year $2006$, aggregated in half an hour intervals. Since the model does not have access to weather information, PV production at neighboring farms is instrumental for obtaining good predictions. Notably, CER-E and PV-US datasets  are at least an order of magnitude larger than the datasets typically used for benchmarking spatiotemporal time series forecasting models. Note that for both PV-US and CER-En the (weighted) adjacency is obtained by applying a thresholded Gaussian kernel to the similarity matrix obtained by considering the geographic distance among the sensors and the correntropy~\cite{liu2007correntropy} among the time series, respectively. We provide further details on the datasets in the supplemental material.

\paragraph{Baselines} We consider the following baselines: 
\begin{enumerate}
    \item \textbf{LSTM}: a single standard gated recurrent neural network~\cite{hochreiter1997long} trained  by sampling window of observations from each node-level time series by disregarding the spatial information;
    \item \textbf{FC-LSTM}: an LSTM processing input sequences as if they were a single high-dimensional multivariate time series;
    \item \textbf{DCRNN}: a recurrent graph network presented in~\cite{li2018diffusion}~--~differently from the original model we use a recurrent encoder followed by a linear readout~(more details in the appendix);
    \item \textbf{Graph WaveNet}: a residual network that alternates temporal and graph convolutions over the graph that is given as input and an adjacency matrix that is learned by the model~\cite{wu2019graph};
    \item \textbf{Gated-GN}: a state-of-the-art time-than-graph~\cite{gao2022equivalence} model introduced in~\cite{satorras2022multivariate} for which we consider two different configurations. The first one~--~indicated as \textbf{FC}~--~uses attention over the full node set to perform spatial propagation, while the second one~--~indicated as \textbf{UG}~--~constrains the attention to edges of the underlying graph. 
    \item \textbf{DynGESN}: the echo state network for dynamical graphs proposed in~\cite{micheli2022discrete}.
\end{enumerate}
For all the baselines, we use, whenever possible, the configuration found in the original papers or in their open-source implementation; in all the other cases we tune hyperparameters on the holdout validation set.

\paragraph{Experimental setup} For the traffic datasets, we replicate the setup used in previous works. In particular, each model is trained to predict the $12$-step-ahead observations. In \gls{model}, the input time series are first encoded by the spatiotemporal encoder, and then the decoder is trained by sampling mini-batches along the temporal dimension, i.e., by sampling $B$ sequences $\gG_{t-W:t}$ of observations. 

For the large-scale datasets, we focus on assessing the scalability of the different architectures rather than maximizing forecasting accuracy. In particular, for both datasets, we consider the first $6$ months of data ($4$ for months for training, $1$ month for validation, and $1$ month for testing). 
The models are trained to predict the next $\{ 00\text{:}30, 07\text{:}30, 11\text{:}00 \}$ hours.
We repeat the experiment in two different settings to test the scalability of the different architectures w.r.t.\ the number of edges. In the first setting, we extract the graph by sparsifying the graph adjacency matrix imposing a maximum of $100$ neighbors for each node, while, in the second case, we do not constrain the density of the adjacency matrix. Tab.~\ref{t:data} reports some details for the considered benchmarks. To assess the performance in terms of scalability, we fix a maximum GPU memory budget of $12$ GB and select the batch size accordingly; if a batch size of $1$ does not fit in $12$ GB, we uniformly subsample edges of the graph to reduce the memory consumption. Differently from the other baselines, in \gls{model} we first preprocess the data to obtain spatiotemporal embeddings and then train the decoder by uniformly sampling the node representations. 
We train each model for $1$ hour, then restore the weights corresponding to the minimum training error and evaluate the forecasts on the test set. The choice of not running validation at each epoch was dictated by the fact that for some of the baselines running a validation epoch would take a large portion of the $1$ hour budget.

The time required to encode the datasets with \gls{model}'s encoder ranges from tens of seconds to $\approx 4$ minutes on an AMD EPYC 7513 processor with $32$ parallel processes. To ensure reproducibility, the time constraint is not imposed as a hard time out; conversely, we measure the time required for the update step of each model on an NVIDIA RTX A5000 GPU and fix the maximum number of updates accordingly. For \gls{model}, the time required to compute node embeddings was considered as part of the training time and the number of updates was appropriately reduced to make the comparison fair. For all the baselines, we keep the same architecture used in the traffic experiment. For \gls{model} we use the same hyperparameters for the decoder, but we reduce the dimension of the embedding~(the value of $K$) so that a preprocessed dataset can fit in a maximum of $\approx 80$ GB of storage. To account for the different temporal scales, we increase the window size for all baselines and increase the number of layers in the ESN~(while keeping the final size of $\overline{\mH}_t$ similar).  Additional details are provided in the supplementary material.

\begin{figure}
    \centering
    \includegraphics[width=\columnwidth]{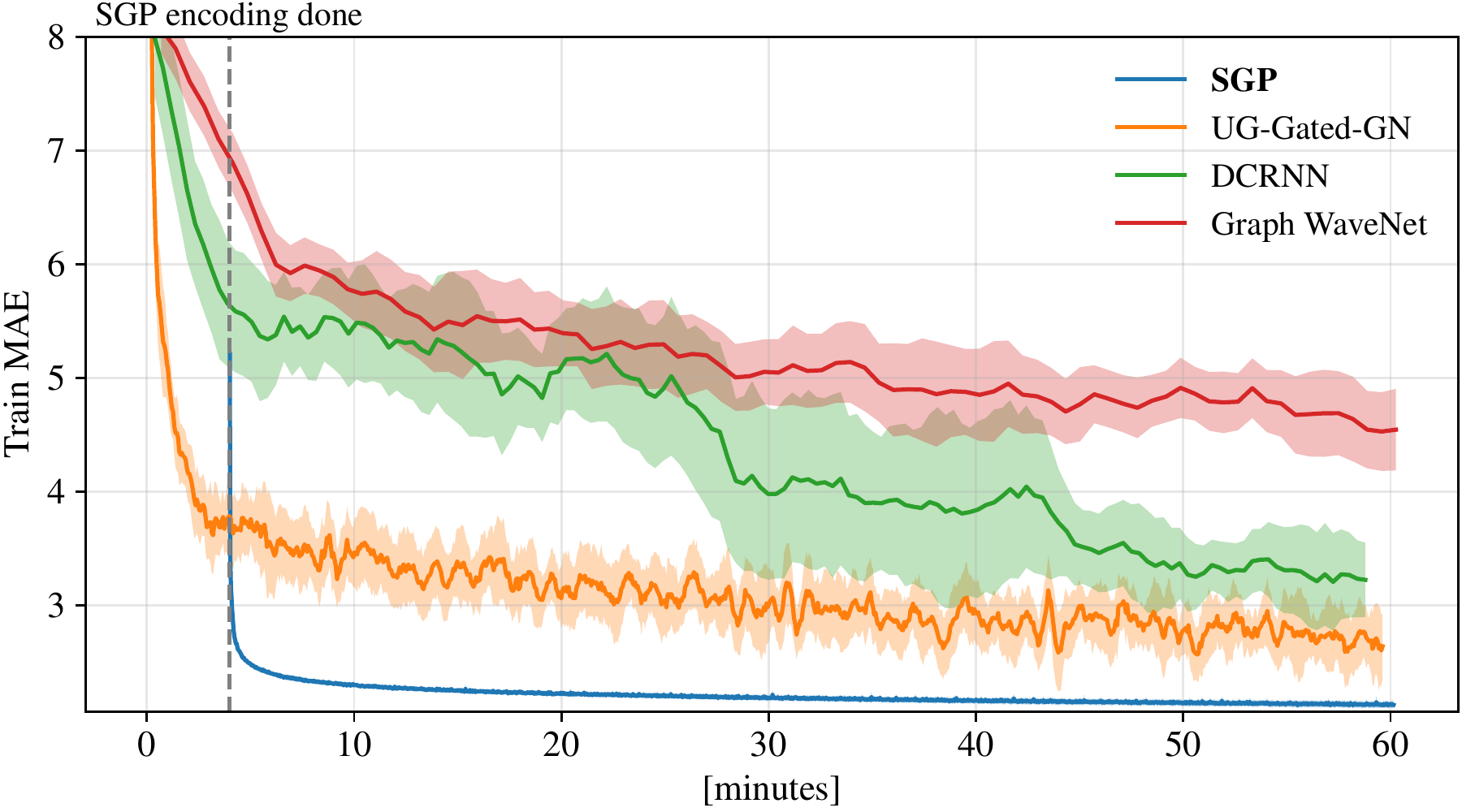}
    \caption{Training curves on PV-US. The plot shows the average $\pm$ the standard deviation of $3$ independent runs. The plotted curves are smoothed with a running average of $8$ steps.}
    \label{fig:training}
\end{figure}

\subsection{Results}

Results for the traffic benchmarks are reported in Tab.~\ref{tab:exptrafficresults}; while the outcomes of the scalability experiments are shown in Tab.~\ref{tab:largescale}. We consider \emph{mean absolute error} (MAE), \emph{mean squared error} (MSE), and \emph{mean absolute percentage error} (MAPE) as evaluation metrics.

\paragraph{Traffic experiment} The purpose of the first experiment is to demonstrate that the proposed method achieves performance comparable to that of the state of the art. In this regard, results in Tab.~\ref{tab:exptrafficresults} show that in all the considered scenarios \gls{model} is always among the best-performing forecasting architectures. The full-attention baseline is the strongest competitor but, however, has time and memory complexities that scale quadratically with the number of nodes. Regarding the other baselines, DCRNN underperforms compared to the other spatiotemporal GNN architectures. DynGESN, the fully randomized architecture, despite being very fast to train, obtains reasonable performance in short-range predictions but falls short over longer forecasting horizons in the considered scenarios. In light of these results, it is worth commenting on the efficiency of \gls{model} compared to the baselines. Approaches like DCRNN and Graph Wavenet, perform graph convolutions whose time and space of complexity is $\bigO(LTE)$, being $E$ the number of edges, $L$ the number of layers ($8$ in Graph Wavenet), and $T$ the time steps.
Such complexity is completely amortized by the preprocessing step in our architecture. Similarly, Gated-GN, while being architecturally much simpler, propagates spatial information by relying on the attention mechanism that is known to scale poorly with the dimensionality of the problem. The next experimental setting shines a light on these shortcomings.

The bottom of Tab.~\ref{tab:exptrafficresults} reports results for the ablation of key elements of the proposed architecture: \textbf{No-Space-Enc.} indicates that the embeddings are built without the spatial propagation step; \textbf{FC-Dec.} consider the case where the structure of the embedding is ignored in the readout and the sparse weight matrix in Eq.~\ref{e:sparse} is replaced by a fully-connected one; \textbf{GC-Dec.} indicates that the spatial propagation is limited to the neighbors of order $K=1$ and, thus, the decoder behaves similarly to a single-layer graph convolutional network. Results clearly show the optimality of the proposed architectural design.

\paragraph{Large-scale experiment} Tab.~\ref{tab:largescale} reports the results of the scalability experiment where we considered only the spatiotemporal GNNs trained by gradient descent. We excluded the full-attention baseline~(FC-Gated-GN) as its $\bigO(N^2)$ complexity prevented scaling to the larger datasets; however, we considered the UG version where attention is restrained to each node's neighborhood. There are several comments that need to be made here. First of all, batch size has a different meaning for our model and the other baselines. In our case, each sample corresponds to a single spatiotemporal (preprocessed) observation; for the other methods, a sample corresponds to a window of observations $\gG_{t-W:t}$ where edges of the graph are eventually subsampled if the memory constraints could not be met otherwise. In both cases, the loss is computed w.r.t.\ all the observations in the batch. The results clearly show that~\gls{model} can be trained efficiently also in resource-constrained settings, with contained GPU memory usage. In particular, the update frequency (batch/s) is up to $2$ order of magnitude higher. Notably, resource utilization at training time remains constant (by construction) in the two considered scenarios, while almost all the baselines require edge subsampling in order to meet the resource constraints. Fig.~\ref{fig:training} shows learning curves for the PV-US dataset, further highlighting the vastly superior efficiency, scalability, and learning stability of~\gls{model}. Finally, results concerning forecasting accuracy show that performance is competitive with the state of the art in all the considered scenarios.

\section{Remarks and conclusion}~\label{sec:conclusion}
We proposed \gls{model}, a scalable architecture for graph-based spatiotemporal time series forecasting. Our approach competes with the state of the art in popular medium-sized benchmark datasets, while greatly improving scalability in large sensor networks. 
While in \gls{model} sampling largely reduces GPU memory usage compared to the other methods, the entire processed sequence can take up a large portion of system memory, depending on the size of the reservoir. Nevertheless, the preprocessing can be distributed, the preprocessed data stored on disk and loaded in batches during training, as customary for large datasets. 
We believe that \gls{model} constitutes an important stepping stone for future research on scalable spatiotemporal forecasting and has the potential of being widely adopted by practitioners. Future work can explore a tighter integration of the spatial and temporal encoding components and assess performance on even larger benchmarks.

\section*{Acknowledgements}

This work was supported by the Swiss National Science Foundation project FNS 204061: \emph{HigherOrder Relations and Dynamics in Graph Neural Networks}. The authors wish to thank the Institute of Computational Science at USI for granting access to computational resources and Nvidia Corporation for the donation of two GPUs.

\bibliography{main}

\appendix

\section*{Appendix}

\input{appendix_content}

\end{document}

%% file: notation.tex
\usepackage{amsmath, amsfonts, bm}

\def\eqref#1{equation~\ref{#1}}

\def\1{\bm{1}}

\def\bigO{\mathcal{O}}

\def\vb{{\bm{b}}}

\def\vh{{\bm{h}}}

\def\vs{{\bm{s}}}

\def\vu{{\bm{u}}}
\def\vv{{\bm{v}}}

\def\vx{{\bm{x}}}

\def\vz{{\bm{z}}}

\def\mTheta{{\bm{\Theta}}}
\def\mA{{\bm{A}}}

\def\mD{{\bm{D}}}

\def\mH{{\bm{H}}}

\def\mS{{\bm{S}}}

\def\mU{{\bm{U}}}
\def\mV{{\bm{V}}}
\def\mW{{\bm{W}}}
\def\mX{{\bm{X}}}

\def\mZ{{\bm{Z}}}

\DeclareMathAlphabet{\mathsfit}{\encodingdefault}{\sfdefault}{m}{sl}
\SetMathAlphabet{\mathsfit}{bold}{\encodingdefault}{\sfdefault}{bx}{n}

\def\gG{{\mathcal{G}}}

\def\sR{{\mathbb{R}}}



%% file: tab_traffic_std.tex
\begin{table*}[t]
\scriptsize
\setlength{\tabcolsep}{2.95pt}
\begin{center}
\resizebox{\textwidth}{!}{%
\begin{tabular}{l | ccc | ccc | ccc | ccc }
\cmidrule[1pt]{2-13}
 \multicolumn{1}{c}{} & \multicolumn{6}{c|}{\bfseries METR-LA} & \multicolumn{6}{c}{\bfseries PEMS-BAY}  \\
\cmidrule[0.8pt]{2-13}
 \multicolumn{1}{c}{} & 15 min & 30 min & 60 min & \multicolumn{3}{c|}{Average} & 15 min & \multicolumn{1}{c}{30 min} & \multicolumn{1}{c|}{60 min} & \multicolumn{3}{c}{Average} \\
\cmidrule[0.5pt]{2-13}
   \multicolumn{1}{c}{} & \tiny MAE & \tiny MAE & \tiny MAE & \tiny MAE & \tiny MSE & \tiny MAPE (\%) & \tiny  MAE & \tiny MAE & \tiny MAE & \tiny MAE & \tiny MSE & \tiny MAPE (\%) \\
 \midrule
 LSTM & 2.99 {\tiny $\pm$ 0.00} & 3.58 {\tiny $\pm$ 0.00} & 4.43 {\tiny $\pm$ 0.01} & 3.58 {\tiny $\pm$ 0.00} & 53.01 {\tiny $\pm$ 0.13} & 10.19 {\tiny $\pm$ 0.05} & 1.39 {\tiny $\pm$ 0.00} & 1.83 {\tiny $\pm$ 0.01} & 2.35 {\tiny $\pm$ 0.01} & 1.79 {\tiny $\pm$ 0.00} & 17.72 {\tiny $\pm$ 0.08} & 4.16 {\tiny $\pm$ 0.05} \\
 FC-LSTM & 3.33 {\tiny $\pm$ 0.01} & 3.43 {\tiny $\pm$ 0.01} & 3.67 {\tiny $\pm$ 0.01} & 3.46 {\tiny $\pm$ 0.01} & 44.85 {\tiny $\pm$ 0.12} & 10.15 {\tiny $\pm$ 0.09} & 2.22 {\tiny $\pm$ 0.01} & 2.25 {\tiny $\pm$ 0.01} & 2.34 {\tiny $\pm$ 0.02} & 2.26 {\tiny $\pm$ 0.01} & 22.31 {\tiny $\pm$ 0.27} & 5.33 {\tiny $\pm$ 0.04} \\
\midrule
DynGESN & 3.27 {\tiny $\pm$ 0.00} & 3.99 {\tiny $\pm$ 0.00} & 5.00 {\tiny $\pm$ 0.00} & 3.98 {\tiny $\pm$ 0.00} & 50.30 {\tiny $\pm$ 0.07} & 11.11 {\tiny $\pm$ 0.01} & 1.57 {\tiny $\pm$ 0.00} & 2.13 {\tiny $\pm$ 0.01} & 2.81 {\tiny $\pm$ 0.02} & 2.09 {\tiny $\pm$ 0.01} & 18.43 {\tiny $\pm$ 0.07} & 4.74 {\tiny $\pm$ 0.01}\\
 DCRNN & 2.82 {\tiny $\pm$ 0.00} & 3.23 {\tiny $\pm$ 0.01} & 3.74 {\tiny $\pm$ 0.01} & 3.20 {\tiny $\pm$ 0.00} & 41.57 {\tiny $\pm$ 0.22} & 8.88 {\tiny $\pm$ 0.05} & 1.36 {\tiny $\pm$ 0.00} & 1.71 {\tiny $\pm$ 0.00} & 2.08 {\tiny $\pm$ 0.01} & 1.66 {\tiny $\pm$ 0.00} & 14.40 {\tiny $\pm$ 0.15} & 3.76 {\tiny $\pm$ 0.01} \\
Graph WaveNet & 2.72 {\tiny $\pm$ 0.01} & 3.10 {\tiny $\pm$ 0.02} & 3.54 {\tiny $\pm$ 0.03} & 3.06 {\tiny $\pm$ 0.02} & 38.22 {\tiny $\pm$ 0.32} & 8.40 {\tiny $\pm$ 0.03} & 1.31 {\tiny $\pm$ 0.00} & 1.64 {\tiny $\pm$ 0.01} & 1.94 {\tiny $\pm$ 0.01} & 1.58 {\tiny $\pm$ 0.00} & 13.12 {\tiny $\pm$ 0.14} & 3.58 {\tiny $\pm$ 0.02} \\
FC-Gated-GN & 2.72 {\tiny $\pm$ 0.01} & \textbf{3.05 {\tiny $\pm$ 0.01}} & \textbf{3.44 {\tiny $\pm$ 0.01}} & 3.01 {\tiny $\pm$ 0.00} & 37.48 {\tiny $\pm$ 0.32} & \textbf{8.27 {\tiny $\pm$ 0.00}} & 1.32 {\tiny $\pm$ 0.00} & 1.63 {\tiny $\pm$ 0.01} & \textbf{1.89 {\tiny $\pm$ 0.01}} & 1.56 {\tiny $\pm$ 0.01} & 12.96 {\tiny $\pm$ 0.11} & 3.51 {\tiny $\pm$ 0.03} \\
UG-Gated-GN & 2.72 {\tiny $\pm$ 0.00} & 3.10 {\tiny $\pm$ 0.00} & 3.54 {\tiny $\pm$ 0.01} & 3.06 {\tiny $\pm$ 0.00} & 38.82 {\tiny $\pm$ 0.08} & 8.40 {\tiny $\pm$ 0.04} & 1.33 {\tiny $\pm$ 0.00} & 1.67 {\tiny $\pm$ 0.01} & 1.99 {\tiny $\pm$ 0.01} & 1.61 {\tiny $\pm$ 0.01} & 13.76 {\tiny $\pm$ 0.14} & 3.59 {\tiny $\pm$ 0.03} \\
\midrule
\textbf{\gls{model}} & \textbf{2.69 {\tiny $\pm$ 0.00}} & \textbf{3.05 {\tiny $\pm$ 0.00}} & \textbf{3.45 {\tiny $\pm$ 0.00}} & \textbf{3.00 {\tiny $\pm$ 0.00}} & \textbf{36.70 {\tiny $\pm$ 0.10}} & \textbf{8.27 {\tiny $\pm$ 0.02}} & \textbf{1.30 {\tiny $\pm$ 0.00}} & \textbf{1.60 {\tiny $\pm$ 0.00}} & \textbf{1.88 {\tiny $\pm$ 0.00}} & \textbf{1.54 {\tiny $\pm$ 0.00}} & \textbf{12.43 {\tiny $\pm$ 0.03}} & \textbf{3.44 {\tiny $\pm$ 0.01}} \\
\midrule
\textit{\textbf{Ablations}} & &  &  &  &  &  &  &  &  &  &  &  \\
--No-Space-Enc. & 2.84 {\tiny $\pm$ 0.00} & 3.26 {\tiny $\pm$ 0.00} & 3.74 {\tiny $\pm$ 0.00} & 3.22 {\tiny $\pm$ 0.00} & 44.55 {\tiny $\pm$ 0.05} & 9.20 {\tiny $\pm$ 0.01} & 1.34 {\tiny $\pm$ 0.00} & 1.68 {\tiny $\pm$ 0.00} & 2.02 {\tiny $\pm$ 0.00} & 1.62 {\tiny $\pm$ 0.00} & 14.14 {\tiny $\pm$ 0.06} & 3.67 {\tiny $\pm$ 0.01}
 \\
--FC-Dec. & 2.76 {\tiny $\pm$ 0.01} & 3.13 {\tiny $\pm$ 0.01} & 3.52 {\tiny $\pm$ 0.02} & 3.08 {\tiny $\pm$ 0.01} & 37.92 {\tiny $\pm$ 0.35} & 8.63 {\tiny $\pm$ 0.11}
 & 1.35 {\tiny $\pm$ 0.01} & 1.67 {\tiny $\pm$ 0.01} & 1.96 {\tiny $\pm$ 0.01} & 1.61 {\tiny $\pm$ 0.01} & 13.04 {\tiny $\pm$ 0.23} & 3.61 {\tiny $\pm$ 0.04}\\
--GC-Dec. & 2.77 {\tiny $\pm$ 0.00} & 3.17 {\tiny $\pm$ 0.00} & 3.63 {\tiny $\pm$ 0.00} & 3.12 {\tiny $\pm$ 0.00} & 40.67 {\tiny $\pm$ 0.06} & 8.74 {\tiny $\pm$ 0.01}
 & 1.32 {\tiny $\pm$ 0.00} & 1.65 {\tiny $\pm$ 0.00} & 1.97 {\tiny $\pm$ 0.00} & 1.59 {\tiny $\pm$ 0.00} & 13.47 {\tiny $\pm$ 0.08} & 3.60 {\tiny $\pm$ 0.01}
\\
\bottomrule
\end{tabular}}
\caption{Results on benchmark traffic datasets (averaged over 3 independent runs). We report MAE, MSE, and MAPE averaged over a one-hour (12 steps) forecasting horizon. We also show MAE for $H\in\{15, 30, 60\}$ minutes time horizons. Bold numbers are within a standard deviation from the best reported average result.}
\label{tab:exptrafficresults}
\end{center}
\end{table*}

%% file: tab_ls.tex
\begin{table*}[t]
\scriptsize
\setlength{\tabcolsep}{2.95pt}
\begin{center}
\resizebox{\textwidth}{!}{%
\begin{tabular}{c | l | ccc | ccc | ccc | ccc }
\cmidrule[1pt]{3-14}
 \multicolumn{2}{c}{} & \multicolumn{6}{c|}{\bfseries PV-US} & \multicolumn{6}{c}{\bfseries CER-En}  \\
\cmidrule[0.8pt]{3-14}
 \multicolumn{2}{c}{} & \multicolumn{3}{c|}{Prediction error (MAE)} & \multicolumn{3}{c|}{Resource utilization} & \multicolumn{3}{c|}{Prediction error (MAE)} & \multicolumn{3}{c}{Resource utilization} \\
\cmidrule[0.5pt]{3-14}
   \multicolumn{2}{c}{} & \tiny 30 mins & \tiny 7 hours 30 mins & \tiny 11 hours & \tiny Batch/s & \tiny Memory & \tiny Batch size & \tiny 30 mins & \tiny 7 hours 30 mins & \tiny 11 hours & \tiny Batch/s & \tiny Memory & \tiny Batch size \\
 \midrule
 \multirow{4}{*}{\rotatebox[origin=r]{90}{100-NN}}
 & DCRNN & 1.39 {\tiny $\pm$ 0.09} & 3.34 {\tiny $\pm$ 0.22} & \textbf{3.54 {\tiny $\pm$ 0.48}} & 2.04 {\tiny $\pm$ 0.01} & 9.63 GB & 2 & 0.22 {\tiny $\pm$ 0.00} & \textbf{0.28 {\tiny $\pm$ 0.00}} & 0.29 {\tiny $\pm$ 0.00} & 1.43 {\tiny $\pm$ 0.02} & 11.10 GB & 2 \\
& Graph WaveNet & 1.45 {\tiny $\pm$ 0.13} & 5.09 {\tiny $\pm$ 0.63} & 5.26 {\tiny $\pm$ 1.34} & 2.01 {\tiny $\pm$ 0.02} & 11.64 GB & 2 & 0.23 {\tiny $\pm$ 0.00} & 0.36 {\tiny $\pm$ 0.01} & 0.36 {\tiny $\pm$ 0.01} & 2.41 {\tiny $\pm$ 0.03} & 8.39 GB & 1 \\
& UG-Gated-GN & 1.33 {\tiny $\pm$ 0.08} & \textbf{2.94 {\tiny $\pm$ 0.05}} & \textbf{3.12 {\tiny $\pm$ 0.14}} & 8.41 {\tiny $\pm$ 0.09} & 11.46 GB & 5 & 0.22 {\tiny $\pm$ 0.00} & \textbf{0.28 {\tiny $\pm$ 0.00}} & \textbf{0.28 {\tiny $\pm$ 0.00}} & 8.21 {\tiny $\pm$ 0.08} & 11.70 GB & 4 \\
\cmidrule[0.5pt]{2-14}
& \textbf{\gls{model}} & \textbf{1.09 {\tiny $\pm$ 0.01}} & \textbf{3.14 {\tiny $\pm$ 0.21}} & \textbf{3.16 {\tiny $\pm$ 0.19}} & \textbf{116.58 {\tiny $\pm$ 8.74}} 
& \textbf{2.21 GB} & 4096 & \textbf{0.21 {\tiny $\pm$ 0.00}} & 0.30 {\tiny $\pm$ 0.00} & 0.31 {\tiny $\pm$ 0.01} & \textbf{117.32 {\tiny $\pm$ 8.36}} 
& \textbf{2.21 GB} & 4096 \\
\midrule

\multirow{4}{*}{\rotatebox[origin=c]{90}{Full}}
& DCRNN & 1.59 {\tiny $\pm$ 0.17} & 4.10 {\tiny $\pm$ 0.27} & 4.93 {\tiny $\pm$ 0.60} & 1.37 {\tiny $\pm$ 0.00} & 11.59 GB & 1$^*$ & 0.23 {\tiny $\pm$ 0.00} & 0.29 {\tiny $\pm$ 0.00} & \textbf{0.29 {\tiny $\pm$ 0.00}} & 1.13 {\tiny $\pm$ 0.01} & 11.10 GB & 1$^*$ \\
& Graph WaveNet & 1.65 {\tiny $\pm$ 0.23} & 6.93 {\tiny $\pm$ 0.58} & 7.93 {\tiny $\pm$ 0.17} & 0.77 {\tiny $\pm$ 0.00} & 11.35 GB & 2 & 0.25 {\tiny $\pm$ 0.01} & 0.38 {\tiny $\pm$ 0.03} & 0.37 {\tiny $\pm$ 0.01} & 1.26 {\tiny $\pm$ 0.01} & 8.58 GB & 1 \\
& UG-Gated-GN & 1.61 {\tiny $\pm$ 0.06} & 3.25 {\tiny $\pm$ 0.04} & \textbf{3.04 {\tiny $\pm$ 0.05}} & 8.83 {\tiny $\pm$ 0.10} & 11.14 GB & 1$^*$ & 0.22 {\tiny $\pm$ 0.00} & \textbf{0.28 {\tiny $\pm$ 0.00}} & \textbf{0.29 {\tiny $\pm$ 0.00}} & 8.77 {\tiny $\pm$ 0.10} & 11.14 GB & 1$^*$ \\
\cmidrule[0.5pt]{2-14}
& \textbf{\gls{model}} & \textbf{1.09 {\tiny $\pm$ 0.00}} & \textbf{3.06 {\tiny $\pm$ 0.11}} & \textbf{3.13 {\tiny $\pm$ 0.13}} 
& \textbf{118.64 {\tiny $\pm$ 8.35}} & \textbf{2.21 GB} & 4096 & \textbf{0.21 {\tiny $\pm$ 0.00}} & 0.30 {\tiny $\pm$ 0.00} & 0.31 {\tiny $\pm$ 0.01} & \textbf{115.85 {\tiny $\pm$ 10.60}} & \textbf{2.21 GB} & 4096 \\
\bottomrule
\end{tabular}}
\caption{Results on large-scale datasets (averaged over at least $3$ independent runs). We report MAE over $H$-step-ahead predictions, $H =$ \{30m, 7h30m, 11h\}, together with timings and memory consumption. $*$ indicates that subsampling was needed to comply with the memory constraints. Bold numbers are within a standard deviation from the best reported average result.}
\label{tab:largescale}
\end{center}
\end{table*}

%% file: tab_data.tex
\begin{table}[ht]
\centering
\small
\begin{tabular}{ l | c c c c }
\toprule
\multicolumn{1}{c|}{Dataset} & \# steps & \# nodes & \# edges & sparsity \\
\midrule
METR-LA & 34272 & 207 & 1515 & 3.54\%\\
PEMS-BAY & 52116 & 325 & 2369 & 2.24\%\\
\midrule
PV-US (100nn) & 8868 & 5016 & 417,199 & 1.66\% \\
CER-En (100nn) & 8868 & 6435 & 639,369 & 1.54\% \\
\midrule
PV-US & 8868 & 5016 & 3,710,008 & 14.75\% \\
CER-En & 8868 & 6435 & 3,186,369 & 7.69\% \\
\bottomrule
\end{tabular}
\caption{Additional information on the considered datasets.}
\label{t:data}
\end{table}

%% file: appendix_content.tex
\section{Detailed experimental settings}

In this appendix, we provide additional details on the experimental settings for the results presented in the paper.

\subsection{Software platform}

The Python~\cite{rossum2009python} code used to run all the computational experiments is available at \url{https://github.com/Graph-Machine-Learning-Group/sgp}.
We relied on the following open-source libraries:
\begin{itemize}
    \item PyTorch~\citep{paszke2019pytorch};
    \item PyTorch Geometric~\citep{fey2019fast};
    \item Torch Spatiotemporal~\cite{Cini_Torch_Spatiotemporal_2022};
    \item PyTorch Lightning~\cite{Falcon_PyTorch_Lightning_2019};
    \item \texttt{numpy}~\citep{harris2020array}.
\end{itemize}

We relied on the Neptune\footnote{\url{https://neptune.ai/}}~\citep{neptune2021neptune} DevOps infrastructure for the logging of the experiments. For all the baselines, we run all the experiments by relying on their open-source implementations.

\subsection{Hardware platform}

Experiments were run on a server equipped with two AMD EPYC 7513 processors and four NVIDIA RTX A5000. Reproducibility of the scalability experiments was ensured by taking timings for the update step of each model and setting the number of updates performed by each model accordingly~(more details in Sec.~\ref{a:training}).

\subsection{Datasets}

All datasets used in our study are open-source or freely available for research purposes. The input graphs are extracted by at first computing a weighted, dense adjacency matrix $\mW$ from (side) spatial information, e.g., the geographic position of the sensors, or by computing a (dis)similarity metric among the time series. The adjacency is then sparsified to obtain $\mA$ by zeroing out values under a certain threshold and, optionally, capping the maximum number of neighbors for each node. For all datasets, the only exogenous variable we consider is the encoding of the time of the day with two sinusoidal functions.

\paragraph{Traffic datasets} Both METR-LA and PEMS-BAY are widely popular benchmarks. We use the same setup of previous works~\cite{wu2019graph} for all the preprocessing steps. As mentioned in Sec.~5, PEMS-BAY contains $6$ months of data from $325$ traffic sensors in the San Francisco Bay Area, while METR-LA contains $4$ months of analogous readings acquired from $207$ detectors in the Los Angeles County Highway~\citep{jagadish2014big}. In both datasets, observations are aggregated at a $5$ minutes time scale.

\paragraph{CER-En} The data from the Irish Commission for Energy Regulation Smart Metering Project~\cite{cer2016cer} contains measurements of the energy consumption aggregated at a $30$ minutes scale in households and small/medium enterprises. The full dataset consists of observations from $6435$ smart meters measuring energy consumption every $30$ minutes. As mentioned in the paper, we use the same preprocessing of~\cite{cini2022filling}, and, in particular, an analogous strategy to extract a graph from the correntropy~\cite{liu2007correntropy} among time series. Note that, differently from~\cite{cini2022filling}, we consider the full sensor network. For all the spatiotemporal GNN baselines, we set the window size to $36$ steps. Access to the dataset can be obtained free of charge by following the information provided at~\url{https://www.ucd.ie/issda/data/commissionforenergyregulationcer}; preprocessing scripts are provided together with the code to reproduce experiments.

\paragraph{PV-US} The PV-US\footnote{\url{https://www.nrel.gov/grid/solar-power-data.html}} dataset~\cite{hummon2012sub} instead consists in a collection of simulated energy production by $\text{5016}$ PV farms for the year $2006$. In the raw datasets, samples are generated every $5$ minute, we aggregate observations at $30$ minutes intervals by taking their mean. A (small) subset of this dataset (often referred to as ``Solar Energy"\footnote{\url{https://github.com/laiguokun/multivariate-time-series-data}}) with only the $137$ PV plants in Alabama state has been used as a multivariate time series forecasting benchmark~\cite{lai2018modeling}.  To obtain an adjacency matrix, we consider the virtual position of the farms in terms of geographic coordinates, and we apply a Gaussian kernel over the pairwise Haversine distances, as described at the beginning of this section. Similarly to the CER-En dataset, we set the window size of the baselines to $36$ steps. The code to download and preprocess the data is available at \url{https://github.com/Graph-Machine-Learning-Group/sgp}.

\subsection{Additional details on SGP architecture}

We implemented the DeepESN encoder following the design principles assessed in previous works~\cite{gallicchio2018design, lukovsevivcius2012practical}. In particular, we decrease the discount factor $\lambda$ progressively at each layer by subtracting $0.1$ from its initial value. We also randomly set $30\%$ of the weights of the networks to $0$ to obtain a sparse reservoir. We use $tanh$ as nonlinear activation function. The recurrent weights are normalized so that the spectral radius of the corresponding matrix is lower than one~\cite{jaeger2001echo}.

For the spatial encoding, we compute the embeddings at the different spatial scales iteratively. Additionally, we also concatenate to the spatiotemporal embedding $\overline{\mS}_t$ the graph-wise average of the temporal embedding $\overline{\mH}_t$ to act as a sort of global attribute~\cite{battaglia2018relational}.

The MLP decoder is implemented as standard feed-forward network with parametrized residual connections between layers~\cite{srivastava2015highway}, SiLU activation function~\cite{hendrycks2016gaussian} and optional Dropout~\cite{srivastava2014dropout} regularization. 

\subsection{Training and evaluation procedure}\label{a:training}

\subsubsection{Traffic}  As previously mentioned, for the traffic datasets we used the same training settings of previous works. For all the baselines we kept the same parameters of previous works whenever possible. For SGP we selected the hyperparameters by performing an initial random search and then manually adjusting the hyperparameters of the reservoir and selecting the best performing configuration on the validation set. In particular, for METR-LA we used a DeepESN with $3$ layers of $32$ units each, an initial decay factor of $0.9$, and a spectral radius of $0.9$. For PEMS-BAY, instead, we used an encoder with a single layer of $128$ units, a decay rate of $0.8$, and a spectral radius of $0.9$. For both datasets, we set $K=4$ and used the bidirectional encoding scheme. In the decoder, for the first layer we used $32$ units for each group in METR-LA and $96$ PEMS-BAY, followed by $2$ fully connected layers of $256$ units each with a dropout rate of $0.3$. The model is trained with early stopping for a maximum of $200$ epochs of $300$ batch each with the Adam optimizer and a multi-step learning rate scheduler.

\subsubsection{Large-scale}
In Tab.~2 of the paper, we report the time required for a single model update~(in terms of batches per second) and GPU memory usage for every considered method.
To ensure a fair assessment, we record the time interval between the beginning of the inference step and the end weights' update for $150$ batches and exclude the first $5$ and last $5$ measurements (that may have overheads). 
We exclude from the computation the overhead introduced -- for every batch -- by the edge subsampling strategy adopted for the scalability of the baselines.

To measure the GPU memory required, we exploit NVIDIA System Management Interface\footnote{\url{https://developer.nvidia.com/nvidia-system-management-interface}}, which provides near real-time GPU usage monitoring.

All the experiments designed to measure time and memory requirements have been run on the same machine on a dedicated reserved GPU. 
We kept the models mostly unchanged w.r.t.\ the traffic experiment. However, we increased the window size to $36$ for the baselines and updated the configuration of the reservoir for SGP to account for the different time scales. In particular, we increased the number of reservoir layers to $8$ and $6$ in PV-US and to in CER-En, respectively, and reduced the number of units accordingly. The difference in the number of layers between the two datasets is motivated by the choice of keeping the size of the preprocessed sequences similar. For this reason, we also set $K=2$ and use the unidirectional encoding to limit the amount of required storage to a maximum $\approx 80$ GB for each dataset.

\subsubsection{Baselines}
For \textbf{LSTM} and \textbf{FC-LSTM} we consider a single-layer LSTM with $128$ units for the temporal embedding and an MLP with one hidden layer with $256$ units and dropout rate of $0.1$. For \textbf{DCRNN}, as reported in~\cite{li2018diffusion}, we set the number of units in the hidden state to $64$ and the order of the diffusion convolution to $K=2$; compared to the original mode, we use a feed-forward readout instead of a recurrent one to enable scalability on the larger benchmarks.
For \textbf{Graph WaveNet} and \textbf{Gated-GN} we use the same hyperparameters and learning rate schedulers reported in the relative papers. We implemented all the baselines in PyTorch and PyTorch Geometric (for graph-based methods) following the open-source implementations provided by the authors. To improve memory and computation efficiency in message-passing layers, we use sparse matrix-matrix multiplications instead of scatter-gather operations whenever possible. We fix the maximum number of training epochs to $300$ to allow all the models to reach convergence, and stop the training if the MAE computed on the validation set does not decrease for $50$ epochs. We evaluate the models using the weights corresponding to the minimum validation MAE.

For \textbf{DynGESN} we set the hyperparameters  of the reservoir to the same ones used for SGP and increase the number of units to approximately match the dimensions of the final embeddings extracted by our method. We trained the readout with Ridge regression by selecting the weight of the L2-regularization term on the validation set.